\documentclass[journal]{IEEEtran}

\usepackage[utf8]{inputenc}
\usepackage[numbers,sort&compress]{natbib}

\usepackage[table]{xcolor}
\usepackage{times}
\usepackage{epsfig}
\usepackage{graphicx}
\usepackage{rotating}
\usepackage{amsmath}
\usepackage{amssymb}
\usepackage{booktabs}
\usepackage{float}
\usepackage{placeins}
\usepackage{multirow}
\usepackage{balance}
\usepackage{multirow}
\usepackage{subfigure}
\usepackage[flushleft]{threeparttable}
\usepackage{textcomp}
\usepackage{url}

\usepackage{enumitem}

\begin{document}
    
        \title{Towards Automated Melanoma Screening: \\ Exploring Transfer Learning Schemes}
    

    \author{
        Afonso Menegola$^\ddagger$, Michel Fornaciali$^\ddagger$, Ramon Pires, Sandra Avila, Eduardo Valle$^\ast$
    \thanks{A. Menegola, M. Fornaciali, S. Avila and E. Valle are with the School of Electrical and Computing Engineering (FEEC), University of Campinas (UNICAMP), Campinas, Brazil (e-mail \{amenegola, michel, sandra, dovalle\}@dca.fee.unicamp.br).}
    \thanks{R. Pires is with the Institute of Computing (IC), University of Campinas (UNICAMP), Campinas, Brazil (e-mail pires.ramon@ic.unicamp.br).}
    
    \thanks{$^\ast${Corresponding author.} | $^\ddagger${These authors contributed equally to this work.}}

    }
    

    \maketitle
    
    \begin{abstract}
        Deep learning is the current bet for image classification. Its greed for huge amounts of annotated data limits its usage in medical imaging context. In this scenario transfer learning appears as a prominent solution. In this report we aim to clarify how transfer learning schemes may influence classification results. We are particularly focused in the automated melanoma screening problem, a case of medical imaging in which transfer learning is still not widely used. We explored transfer with and without fine-tuning, sequential transfers and usage of pre-trained models in general and specific datasets. Although some issues remain open, our findings may drive future researches. 
    \end{abstract}
    

    %

    \section{Introduction}

Melanoma is the leading cause of deaths due to skin cancer. Its prognosis is very good when detected early, but deteriorates rapidly as the disease progresses. Therefore, early diagnosis is critical and screening --- the search for new cases --- must be a continuous process. 

Image processing can help melanoma screening programs, differentiating malignant from benign skin lesions. Nowadays, image classification has been mostly done through deep learning techniques. Unfortunately, it is not common to find huge amounts of medical data enabling medical computer vision with deep learning. So researchers usually employ \textbf{transfer learning} techniques in order to deal with the lack of enough annotated images. 

In this report we aim to clarify how transfer schemes may influence the final results of automated melanoma screening. The main aspects under investigation are: (1) if (and how) consecutive transfer schemes --- specializing the classification tasks along the pipeline --- improve the results; (2) if transfer learning done between similar datasets/tasks improve the results; and (3) how much fine tuning improves results for small datasets. 

Our main contributions are not in understanding fine level details about transfer learning (e.g. parametrization), but to clarify how transfer schemes should be organized to improve final results. Although we focus in the automated melanoma screening problem, we understand that the main findings may generalize to other specific-context datasets. 

A complete review of transfer learning and automated melanoma screening is beyond the scope of this paper. In order to know the motivation, technical implementation details and related works in these fields, please, see our last reports \cite{fornaciali2016towards, fornaciali2014statistical}.

    \section{Approaches} 
\label{sec:proposed}

In visual tasks, the low-level layers of a deep neural network (DNN) tend to be fairly general; specialization increases as we move up in the network~\cite{Yosinski2014}.

Therefore, transfer learning is commonly used on a straightforward way: you just need to freeze the weights of a pre-trained original DNN up to a chosen layer, replacing and retraining the other layers for the new task or even plugging an SVM classifier on the top layer (or any other classifier if you want to). This approach is called \textit{vanilla transfer learning without fine-tuning}. 

In \textit{transfer learning with fine-tuning} we update the network weights to adjust a original model for the desired task, improving classification results.

In this report, we forgo comparisons with the state-of-the-art, in order to focus on these questions: what is a good transfer learning scheme? Which original models work best for transfer? What is the relative impact of the original model choice versus the use of fine-tuning? 

Let's investigate two approaches of transfer: \textbf{simple} and \textbf{combined}. Among simple transfer learning, let's analyse the results \textit{with} and \textit{without} fine tuning, selecting a model trained in a huge \textit{general-context dataset} (e.g. ImageNet) or a model trained in a smaller \textit{specific-domain dataset} (e.g. images of medical domain). The combined approach concatenates two sequential transfer steps: the first starting with a powerful general-context dataset, transferring the knowledge to a specific-domain dataset used to refine the model. The refined model is then used to transfer knowledge again, now to the target dataset (in our case, melanoma images). 

As baseline, we will also use a DNN trained from scratch. That setup does not involve transfer learning: the model is generated from and tested only with melanoma images. 

All approaches (from scratch, and with transfer learning) use the same DNN architecture: we adopt the \textit{VGG-M} model proposed by Chatfield~et~al.~\citep{Chatfield2014}. When we use transfer learning, we copy all layers but the last from the pre-trained VGG-M model and adapt the final layer for the target task. When we fine-tune, we exchange the model output layer for a softmax output layer with two or three classes, according to the experiment, and train that complete neural model as usual, backpropagating the errors and updating the weights throughout the network. However, in all networks, including the fine-tuned ones, we ignore the output layer and employ an SVM classifier to make the decision, using the next-to-last layer output as features (Section~\ref{sec:results}). We describe the overall procedures next. 

\subsection{Simple transfer learning WITHOUT fine tuning}\label{sec:sem_fine_tuning}

\begin{enumerate}
    \item The VGG-M network model is defined using Lasagne library\footnote{Lasagne library: https://lasagne.readthedocs.io/en/latest/};
    \item We load the weights of a pre-trained network. In our approaches we have two options: 
    \begin{enumerate}
        \item ImageNet: we used a pre-trained MatConvNet VGG-M model\footnote{MatConvNet: http://www.vlfeat.org/matconvnet/pretrained/}, trained in the ImageNet dataset. Since the model available is a .mat file, it was necessary to transcript the weights to a Lasagne-readable format (we distribute the script that implements this transcription); 
	    \item Medical domain: the VGG-M model is trained from scratch using a dataset of medical images. See specific details at \ref{sec:scratch}; 
    \end{enumerate}
    \item To match the input images to the size required by VGG-M, we resize all images to $224\times 224$ pixels, using Pillow/ANTIALIAS\footnote{Pillow Imaging Library (version 2.3.0): https://pillow.readthedocs.io/}, distorting the aspect ratio to fit when needed;
    \item As a centering step, all input images are subtracted by the dataset mean used to train the model. For the ImageNet case, the mean is available in the .mat file whilst the mean from the medical dataset was calculated inside the code. These centralized images fed the pre-trained model; 
    \item We use the weights of the pre-trained network to extract the outputs of Group 7/Layer 19, which are vectors of $4096$ dimensions that describe the input images;
    \item We $\ell_2$-normalize those vectors, which will be mid-level features for the classifying step; 
    \item We separate 10\% of the training set for validation; 
    \item We use the mid-level features from the remaining of the training set to feed a linear SVM classifier. We choose the Sklearn implementation\footnote{Sklearn: http://scikit-learn.org/stable/}. We performed a grid search exploring the margin hardness $C\in \{10^c:c\in[-4:3]\}$, seeking through the use of internal cross-validation for the best SVM classifier that minimizes the F-score (since this score can be used for all experimental designs (see Section~\ref{sec:designs}); 
    \item We incorporate the validation set into the training set and create a \textit{final SVM model} using the best $C$ of the grid search; 
    \item Finally, we use that ``final SVM model'' and the mid-level features from the melanoma testing set to obtain the reported mean Average Precision scores.
\end{enumerate}

\subsection{Simple transfer learning WITH fine tuning}\label{sec:com_fine_tuning}    

\begin{enumerate}
    \item We do steps 1) to 4) of \ref{sec:sem_fine_tuning}; 
        
    \item We augment the training set generating new perturbed images in order to balance the classes. The perturbations are: zoom, rotation, shear, translation, flipping and stretching transformations. We apply those transformations in each image, with parameters chosen at random. We include the new generated images in the training set, while images of the most favored class are excluded at random until the classes are balanced; 
    
        
    \item We perform a training step with 200 epochs over the loaded model with a learning rate schedule, starting with $10^{-3}$, and reducing to $10^{-4}$ at epoch 100, and then to $10^{-5}$ at epoch 150. As the model is being fine-tuned, we save the weights that minimizes the validation loss; 
        
    \item Then, we do the steps 5) to 10) of \ref{sec:sem_fine_tuning}, but in step 5) we use the weights with lowest validation loss to extract the outputs of Group 7/Layer 19;
        
\end{enumerate}

\subsection{Combined transfer learning}\label{sec:combined} 
The combined transfer learning is a sequence of two consecutive simple transfers. First we load the ImageNet model and fine tune it to a smaller medical domain dataset. Then we use that fine tuned model to perform a second fine tuning step, now over the melanoma dataset.


\subsection{Training a DNN from scratch}\label{sec:scratch}  
As mentioned before, our baseline is a DNN model trained from scratch using melanoma images. The steps are similar of transfer learning with fine tuning (\ref{sec:com_fine_tuning}), but with small differences: 

\begin{itemize}
    \item We initialize the network with random weights; 
    \item The mean used to center the images are now calculated over the training set of the melanoma dataset; 
\end{itemize}

    \section{Experimental Details} \label{sec:experimental}

\subsection{Datasets}
\label{sec:datasets}

\subsubsection{Melanoma}
We used the dataset of the Interactive Atlas of Dermoscopy~\citep{argenziano2002dermoscopy}. This Atlas is a multimedia guide (Booklet + CD-ROM) designed for training medical personnel to diagnose skin lesions. The CD-ROM contains 1000+ clinical cases, each with at least two images of the lesion: close-up clinical image, and dermoscopic image. Most images are 768 pixels wide $\times$ 512 high. Besides the images, each case is composed by clinical data, histopathological results, diagnosis, and level of difficulty. The latter measures how difficult (low, medium and high) the case is considered to diagnose by a trained human. The diagnoses include, besides melanoma (several subtypes), basal cell carcinoma, blue nevus, Clark's nevus, combined nevus, congenital nevus, dermal nevus, dermatofibroma, lentigo, melanosis, recurrent nevus, Reed nevus, seborrheic keratosis, and vascular lesion. There is also a small number of cases classified simply as `miscelaneous'. 

\subsubsection{Other datasets}
The following datasets were only used train original models for transfer learning. 

\begin{itemize}
    \item Diabetic Retinopathy: the other specific-domain dataset used on our experiments is the training set of the the Kaggle Challenge for Diabetic Retinopathy Detection\footnote{https://www.kaggle.com/c/diabetic-retinopathy-detection/data}. This dataset is composed by more than 35,000 high-resolution retina images taken under a variety of imaging conditions. More information can be found at the challenge website; 
    \item ImageNet: as general-context dataset, we employed the ILSVRC-2012 challenge dataset, containing about 1.2M training images of 1,000 object categories from ImageNet~\citep{deng2009imagenet}. As mentioned before, we did not train this dataset from scratch, but used the MatConvNet VGG-M pretrained model, just converting the .mat file (containing the complete description of the network and the all layer weights pre-trained for the ImageNet task) for our framework. 
\end{itemize}

\subsection{Experimental Designs}
\label{sec:designs}

We investigated three protocols, trying to identify if (and how) label variations can impact the method. The protocols are: 

\begin{enumerate}
    \item Malignant vs. Benign lesions: melanomas and basal cell carcinomas were considered positive cases and all other diagnoses were negative cases; 
    \item Melanoma vs. Benign lesions: melanomas were positive cases while all other diagnoses were negative ones, removing basal cell carcinomas;
    \item Basal cell carcinoma vs. Melanoma vs. Benign lesions: here we have three classes, one for melanoma, other for basal cell carcinomas and all other diagnoses were classified under a single third label. 
\end{enumerate}

For each protocol we employed 5$\times$2-fold cross-validation, that is, the data was `randomly' splitted in two groups: A and B. We trained in group A and tested in group B. Then, we reverted the protocol: we trained in B and tested in A. We have done 5 semi-random splits, making an effort to balance each group according to the diagnosis of the case (that is, almost 50\% of the cases of each diagnosis for each group A and B). 


We only used dermoscopic images, removing the ones with acral lesions. We used images of all difficulties: low, medium and high. We do not removed images with hair, dots, rulers and other signs not belonged by the lesions. Some images contained a black ``frame'' around the picture, which we removed, cropping by hand. 

We used the mean Average Precision (mAP) for protocols (1) and (2). For protocol (3) (with three classes) we employed the \textit{macro} mean Average Precision. The reference implementation adopted for both metrics were the ones used at PASCAL VOC 2007\footnote{PASCAL VOC 2007 Development Kit: http://host.robots.ox.ac.uk/pascal/VOC/voc2012/\#devkit}.

\subsection{Experiments}
\label{sec:experiments}

We performed six experiments with each protocol, always using the VGG-M DNN architecture~\citep{Chatfield2014}. The feature vectors were extracted using the 19th layer and the classification was done using SVM. The experiments are: 

\begin{enumerate}[label=\Alph*]
\item Training and testing a network from scratch with melanoma images;
\item Training a network from scratch with diabetic retinopathy images, transfer learning for melanoma \textbf{without} fine tuning; 
\item Training a network from scratch with diabetic retinopathy images, transfer learning for melanoma \textbf{with} fine tuning; 
\item Uploading a pretrained network with ImageNet images, transfer learning for melanoma \textbf{without} fine tuning; 
\item Uploading a pretrained network with ImageNet images, transfer learning for melanoma \textbf{with} fine tuning; 
\item Uploading a pretrained network with ImageNet images, transfer learning for diabetic retinopathy \textbf{with} fine tuning then transferring again for melanoma images \textbf{with} fine tuning; 
\end{enumerate}

A reference implementation for our approaches is available in the repository linked at our website\footnote{To find the source code of this paper, visit our website: https://sites.google.com/site/robustmelanomascreening}.




    \section{Results}
\label{sec:results}

We show our results in Table~\ref{results}. The main findings are: 

\begin{itemize}
    
    \item As expected, training a DNN from scratch is not essentially better than performing transfer learning with fine-tuning. Moreover, training a DNN from scratch is time consuming~\citep{tajbakhsh2016convolutional};

    \item Performing fine-tuning improves the classification results. It's true for both transfer learning from specific-domain dataset and from general-context. This result was already expected, according to the literature~\citep{Yosinski2014}; 
 
    \item Surprisingly, transfer learning between less related tasks (from ImageNet to melanoma) performed better than tasks in the same domain (medical images, from retina to melanoma). This result is consistent among all protocols, independently from employing fine-tuning or not. 
 
    \item Even more surprising, the combined transfer performed worse than the simple transfer. Besides that: the combined transfer had a similar result of the transfer from retina dataset, as if the whole knowledge learned from ImageNet was ``erased'' by the specific-domain dataset on its fine-tuning step; 
    
    \item Melanoma and basal cell carcinomas are two types of skin cancers. When both types are grouped under the \textit{positive} class (first line), the classification improves. Maybe this occur because the unbalancing of \textit{positive} and \textit{negative} classes is smaller; 
    
    \item Removing basal cell carcinomas from the experiments diminish the classification results, maybe because the models have fewer images to learn the differences between them (second line, regarding the other ones);
    
\end{itemize}

\begin{table*}[]
\centering
\caption{Main results for automated melanoma screening, expressed in (average $\pm$ standard deviation). FT: fine tuning.}
\label{results}
\begin{tabular}{|c|c|c|c|c|c|c|}
\hline
\multirow{3}{*}{\textbf{\begin{tabular}[c]{@{}c@{}}Experimental Designs \\ (Protocols)\end{tabular}}} & \multicolumn{6}{c|}{\textbf{Experiments (mAP (\%))}}                                                                                                                                         \\ \cline{2-7} 
                                                                                                      & \multirow{2}{*}{\textbf{Baseline}} & \multicolumn{2}{c|}{\textbf{Transfer from retina}} & \multicolumn{2}{c|}{\textbf{Transfer from ImageNet}} & \multirow{2}{*}{\textbf{Combined transfer}} \\ \cline{3-6}
                                                                                                      &                                    & \textbf{no FT}          & \textbf{with FT}         & \textbf{no FT}           & \textbf{with FT}          &                                             \\ \hline
\textit{Malignant vs. Benign}                                                                         & 55.4 $\pm$ 2.5                        & 49.3 $\pm$ 1.2              & 57.1 $\pm$ 4.0               & 69.8 $\pm$ 2.4               & 73.0 $\pm$ 2.9                & 57.4 $\pm$ 1.1                                  \\ \hline
\textit{Melanoma vs. Benign}                                                                          & 53.3 $\pm$ 4.3                         & 47.5 $\pm$ 1.9              & 53.1 $\pm$ 2.7               & 60.5 $\pm$ 3.1               & -                         & 54.4 $\pm$ 3.9                                  \\ \hline
\textit{Basal cell carcinoma vs. Melanoma vs. Benign lesions}                                         & 53.1 $\pm$ 1.6                         & 48.9 $\pm$ 1.6              & 51.6 $\pm$ 2.6               & 61.6 $\pm$ 2.9               & -                         & 52.9 $\pm$ 1.3                                  \\ \hline
\end{tabular}
\end{table*}

As we mentioned before, although the fine-tuning processes occur in a completely neural network pipeline, we choose to show the final results using an SVM classifier in order to enable fair comparisons with past experiments~\citep{fornaciali2016towards}: the results of transfer learning from ImageNet without fine tuning are the most comparable to the ones reported in our previous publication (regarding small differences at experimental designs). Since the results are very similar --- with both codes employing VGG-M DNN + SVM classifier --- we infer that the code used in this paper is correct. 

Some explanations for transfer from retina dataset be worse than transfer from ImageNet are (a) that the the last model was much more optimized than the first one, (b) the ImageNet dataset is much bigger than the retina dataset and also (c) that the retina model was created with unbalanced training set. 

Maybe the combined transfer did not perform as well as expected because diabetic retinopathy may be easier to diagnose than melanoma. So in the first transfer step the network did not learn to be as specialized as needed to ``see'' details/differences on skin lesion images. This can also justify why transfer from retina is worse than transfer from ImageNet.

    \section{Conclusions}
\label{sec:conclusions}

In this report we investigate how different transfer learning schemes influence automated melanoma classification results. We evaluated transfer learning from a general-context dataset (ImageNet) and from a specific-domain dataset (diabetic retinopathy). We also investigated if sequential transfer steps improve the final classification result. 

We show results consistent with the literature regarding training a DNN from scratch and differences between doing fine-tuning or not. We conclude that general findings of deep learning state-of-the-art are also applicable for automated melanoma screening literature, thus guiding future research. 

Although we expected that transfer learning from related tasks (in our case, \textit{from} and \textit{to} medical domain datasets) could lead to better results, it was not observed. Some conditions that may had influenced the results are the dataset sizes, parametrization used for training the models and quality of the datasets (in terms of annotations, standardization and image acquisition processes). In this case, further investigation is needed. 

We also conclude that the experimental design is sensitive to the image annotation, that is, small changes in the fold assembling can cause huge impacts in the final results. This finding is particularly important and will be discussed in future experiments.

    \section*{Acknowledgements}
    We gratefully acknowledge the support of NVIDIA Corporation with the donation of the Tesla K40 GPU used for this research. We are also grateful to Prof. Dr. M. Emre Celebi for kindly providing the machine-readable matadata of The Interactive Atlas of Dermoscopy. A. Menegola is funded by CNPq; S. Avila is funded by PNPD/CAPES; R. Pires is funded by CAPES; M. Fornaciali, R. Pires and E. Valle are partially funded by Google Research Awards for Latin America 2016.


    \balance
    \bibliographystyle{IEEEtranN}
    \bibliography{references}
    
\end{document}